\pdfoutput=1

\documentclass[11pt]{article}

\usepackage[]{acl}
\usepackage{multirow}

\usepackage{times}
\usepackage{latexsym}
\usepackage{array}
\usepackage{enumitem}

\usepackage{booktabs}
\usepackage{multicol}
\usepackage{blindtext}
\usepackage{amssymb}
\usepackage[textsize=scriptsize]{todonotes}

\usepackage{graphicx} 
\graphicspath{ {./figs/} }

\usepackage[T1]{fontenc}

\usepackage[utf8]{inputenc}

\usepackage{microtype}

\newcommand{\fy}[1]{\textcolor{blue}{\textbf{fangyuan:} #1}}
\newcommand{\ec}[1]{\textcolor{cyan}{\textbf{eunsol:} #1}}

\usepackage{caption}
\usepackage{subcaption}
\usepackage{pgfplots}
\usetikzlibrary{patterns}
\usepackage{filecontents}
\usepackage{algorithmic,algorithm}
\usepackage{csvsimple}
\usepackage{amsmath}

\definecolor{blue}{HTML}{4285f4}
\definecolor{lightblue}{HTML}{c9daf8}
\definecolor{darkblue}{HTML}{1c4587}
\definecolor{green}{HTML}{34a853}
\definecolor{lightgreen}{HTML}{d9ead3}
\definecolor{orange}{HTML}{ff9900}
\definecolor{lightorange}{HTML}{fce5cd}
\definecolor{lightred}{HTML}{e06666}
\definecolor{purple}{HTML}{9900ff}
\definecolor{lightpurple}{HTML}{b4a7d6}
\definecolor{gray}{HTML}{cccccc}

\definecolor{sqcolor}{HTML}{4285f4} 
\definecolor{nqcolor}{HTML}{0b5394} 
\definecolor{hpcolor}{HTML}{9fc5e8} 
\definecolor{trvcolor}{HTML}{ff9900} 
\definecolor{seacolor}{HTML}{f9cb9c} 
\definecolor{newscolor}{HTML}{cc4125} 

\title{How Do We Answer Complex Questions:\\
Discourse Structure of Long-form Answers}

\author{Fangyuan Xu$^{\diamondsuit}$~~  Junyi Jessy Li$^{\heartsuit}$ ~~ Eunsol Choi$^{\diamondsuit}$ \\
$^\diamondsuit$Department of Computer Science \\
$^\heartsuit$Department of Linguistics \\
 The University of Texas at Austin \\
 \hspace{0.5em} {\texttt{\{fangyuan, jessy, eunsol\}@utexas.edu}} \\}

\begin{document}

\maketitle

\begin{abstract}
Long-form answers, consisting of multiple sentences, can provide nuanced and comprehensive answers to a broader set of questions. To better understand this complex and understudied task, we study the functional structure of long-form answers collected from three datasets, ELI5~\cite{Fan2019ELI5LF}, WebGPT~\cite{nakano2021webgpt} and Natural Questions~\cite{kwiatkowski2019natural}. Our main goal is to understand how humans organize information to craft complex answers. We develop an ontology of six sentence-level functional roles for long-form answers, and annotate 3.9k sentences in 640 answer paragraphs. Different answer collection methods manifest in different discourse structures. We further analyze model-generated answers -- finding that annotators agree less with each other when annotating model-generated answers compared to annotating human-written answers. Our annotated data enables training a strong classifier that can be used for automatic analysis. {We hope our work can inspire future research on discourse-level modeling and evaluation of long-form QA systems.}\footnote{Our data, code and datasheet are available at \url{https://github.com/utcsnlp/lfqa_discourse}.}
\end{abstract}

\section{Introduction}

While many information seeking questions can be answered by a short text span, requiring a short span answer significantly limits the types of questions that can be addressed as well as the extent of information that can be conveyed. 
Recent work~\cite{Fan2019ELI5LF,Krishna2021HurdlesTP,nakano2021webgpt} explored long-form answers, where answers are free-form texts consisting of multiple sentences.
Such long-form answers provide flexible space where the answerer can provide a nuanced answer, incorporating their confidence and sources of their knowledge. 
Thus the answer sentences form a \emph{discourse} where the answerers provide information, hedge, explain, provide examples, point to other sources, and more; these elements need to be structured and organized coherently.

We take a linguistically informed approach to understand the structure of long-form answers, designing six communicative \emph{functions} of sentences in long-form answers (which we call \emph{\bf roles}).\footnote{Functional structures have been studied in various other domains (discussed in Sections~\ref{sec:role_ontology} and~\ref{sec:related}).} Our framework combines functional structures with the notion of information salience by designating a role for sentences that convey the main message of an answer. 
Other roles include signaling the organization of the answer, directly answering the question, giving an example, providing background information, and so on. About a half of the sentences in long-form answers we study serve roles other than providing an answer to the question.

\begin{figure*}
    \centering
    \includegraphics[width=1\textwidth]{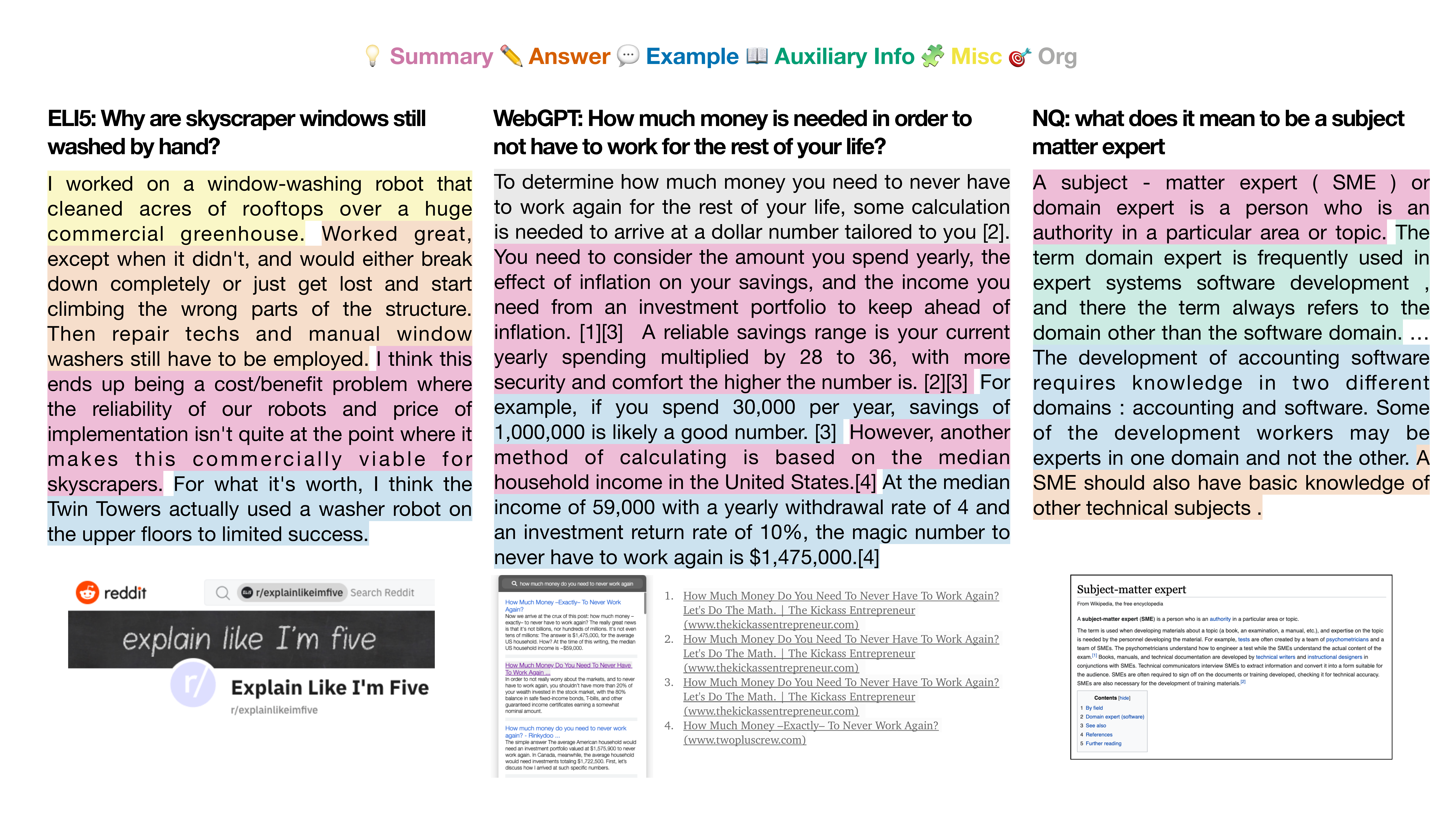}
    \caption{Long-form answers from ELI5, WebGPT and NQ dataset. Each sentence in the answer is annotated with a sentence-level functional role from our ontology, described in Section \ref{sec:role_ontology}.}
    \label{fig:intro_1}
\end{figure*}

We collect discourse annotations on three long-form question answering (LFQA) datasets, ELI5~\cite{Fan2019ELI5LF}, WebGPT \cite{nakano2021webgpt} and Natural Questions (NQ) ~\cite{kwiatkowski2019natural}. Figure~\ref{fig:intro_1} contains an example annotation on each dataset. 
While all three contain paragraph-length answers needed for complex queries, they are collected in distinct manners -- answers in ELI5 are written by Reddit users; answers in WebGPT are written by annotators who searched documents on a web interface and heavily quoted those documents to form an answer, and answers in NQ are pre-existing paragraphs from Wikipedia corpus. We collect three-way annotations for 3.9k sentences ($\sim$700 question-answer pairs across three datasets). We also annotate a small number of model-generated answers from a recent long-form question answering (LFQA) system \cite{Krishna2021HurdlesTP} and provide rich analysis of their discourse structure.



In all three datasets, we observe appearance of most proposed functional roles, but with different proportions. Answers in ELI5 contains more examples and elaborations, while answers extracted from Wikipedia passages (NQ) contain more auxiliary information. Analyzing a subset of ELI5 and WebGPT, we also identify a big gap in lexical overlap between long-form answer and evidence passages across all functional roles. Lastly, we found that human agreement of the discourse roles of model-generated answers are much lower than human-written ones, reflecting the difficulty for humans to process model-generated answers.


With the data collected, we present a competitive role classifier, which performs on par with human when trained with our annotated data and can be used for automatic discourse analysis. We further envision using functional roles for controllable long-form generations, concise answer generation, and improved evaluation metrics for LFQA. 

\section{Defining Answer Discourse Structure}\label{sec:role_ontology}
We study the discourse structure of long-form answers based on \textit{functional roles} of sentences in the paragraph. Functional structures characterize the communicative role a linguistic unit plays; as such, they vary across genres as the goals of communication also vary. In scientific or technical articles, these roles can be \emph{background, method, findings}~\cite{kircz1991rhetorical,liddy1991discourse,mizuta2006zone}, while in news, they can be \emph{main event} or \emph{anecdotes}~\cite{van2013news,choubey2020discourse}.

These structures are related to, though distinct from, coherence discourse structures~\cite{hobbs1985coherence}. The latter characterizes how each unit (e.g., adjacent clauses or sentences) \emph{relates} to others through semantic relations such as temporal, causal, etc.; such structures can be trees that hierarchically relate adjacent units~\cite{mann1988rhetorical} or graphs~\cite{lascarides2008segmented}. In contrast, functional roles describe how information is organized to serve the communication goal, in our case, providing the answer. 

We developed our ontology by examining long-form answers in online community forums (subreddit \textit{Explain
Like I’m Five} (ELI5)) and Wikipedia passages, hence answers derived from different domains (e.g., textbooks) can contain roles beyond our ontology. We describe our six sentence-level discourse roles for long-form answers here:

\paragraph{Answer-Summary (Sum), Answer (Ans).} An answer sentence directly addresses the question. 
Here we distinguish between the the main content of the answer (henceforth \textit{answer summary}) vs.\ sentences which explain or elaborate on the summary.
The summaries play a more salient role than non-summary answer sentences, and can often suffice by themselves as the answer to the question. This is akin to argumentation structure that hierarchically arranges main claims and supporting arguments~\cite{peldszus2013argument}, and news structure that differentiates between main vs.\  supporting events~\cite{van2013news}.

\paragraph{Organizational sentences (Org.)}
Rather than conveying information of the answer, the major role of an organizational sentence is to inform the reader how the answer will be structured. We found two main types of such sentences; the first signals an upcoming set of items of parallel importance:
\advance\leftmargini -1em
\begin{quote}
\small
    \textbf{[A]}: \textbf{There are a few reasons candidates with ``no chance" to win keep running}. 1) They enjoy campaigning[...]
\end{quote}
The other type indicates that part of the answer is upcoming amidst an established flow; in the example below, the answerer used a hypophora:
\begin{quote}
\small
    \textbf{[A]}: It might actually be a mosquito bite. I find the odd mosquito in my house in the winter from time to time, and I'm in Canada.[...] \textbf{So why does it happen more often when you shower?} It's largely because [...]
\end{quote}
\paragraph{Examples (Ex.)} Often people provide examples in answers; these are linguistically distinct from other answer sentences in the sense that they are more specific towards a particular entity, concept, or situation. This pattern of language specificity can also be found in example-related discourse relations~\cite{louis2011automatic,li2015fast}, or through entity instantiation~\cite{mackinlay2011modelling}:
\begin{quote}
\small
    \textbf{[Q]}: What is it about electricity that kills you?\\
    \textbf{[A]}: [...] \textbf{For example, static electricity consists of tens of thousands of volts, but basically no amps.} [...]
\end{quote}
We found that examples in human answers are often not signaled explicitly, and often contain hypothetical situations:
\begin{quote}
\small
    \textbf{[Q]}: Were major news outlets established with political bias or was it formed over time?\\
    \textbf{[A]}: [...]This is impossible due to the problem of ``anchoring.'' \textbf{Consider a world where people on the right want the tax rate to be 1\% lower and people on the left want the tax rate to be 1\% higher}[...]
\end{quote}

\paragraph{Auxiliary information (Aux.)} These sentences provide information that are related to what is discussed in the answer, but not asked in the question. It could be background knowledge that the answerer deemed necessary or helpful, e.g.,
\begin{quote}
\small
    \textbf{[Q]}: Why is it better to use cloning software instead of just copying and pasting the entire drive?\\
    \textbf{[A]}: \textbf{When you install an operating system, it sets up what's called a master file table, which [...] 
    are important for the OS to work properly.} [...] Simply copy-pasting files doesn't copy either of these, meaning if you want to back up an OS installation you should clone the disk instead. 
\end{quote}
or related content that extends the question, e.g.,
\begin{quote}
\small
    \textbf{[Q]}: what is the difference between mandi and kabsa?\\
    \textbf{[A]}: [...] A popular way of preparing meat is called mandi. [...] \textbf{Another way of preparing and serving meat for kabsa is mathbi , where seasoned meat is grilled on flat stones that are placed on top of burning embers.}
\end{quote}
Notably, the removal of auxiliary information would still leave the answer itself intact.

\paragraph{Miscellaneous (Misc.)} 
We observe various roles that, although less frequent, show up consistently in human answers. We group them into a \emph{miscellaneous} role and list them below.

(a) Some sentences specify the limitation of the answer by narrowing down the scope of the answer to an open-ended question.
\begin{quote}
\small
    \textbf{[Q]}: Why are there such drastic differences in salaries between different countries?\\
    \textbf{[A]}: I’m going to avoid discussing service industries, because[...] \textbf{I’m mostly talking tech.} [...]
\end{quote}
(b) Some sentences state where the answer came from and thus put the answer into context.
\begin{quote}
\small
     \textbf{[Q]}: Why Does a thermostat require the user to switch between heat and cool modes, as opposed to just setting the desired temperature?\\
    \textbf{[A]}: \textbf{The person who installed my heat pump (which has all three modes) explained this to me.} [...]  
\end{quote}
(c) Some sentences point to other resources that might contain the answers.
\begin{quote}
\small
    \textbf{[Q]}: Why did Catholicism embrace celibacy and why did Protestantism reject it?\\
    \textbf{[A]}: \textbf{\slash r\slash askhistorians has a few excellent discussions about this.} [...]
\end{quote}
(d) Answerers also express sentiment towards other responses or the question itself.
\begin{quote}
\small
    \textbf{[Q]}: Why did Catholicism embrace celibacy and why did Protestantism reject it?\\
    \textbf{[A]}: \textbf{Good God, the amount of misinformation upvoted is hurting.} [...]
\end{quote}

\begin{quote}
\small
    \textbf{[Q]}: Could you Explain Schrödinger's Cat to me LI5?\\
    \textbf{[A]}: [...] \textbf{It's a pretty cool thought experiment, but it doesn't mean too much in our everyday lives.}
\end{quote}

{As our ontology does not provide an exhaustive list of the functional roles, we instructed our annotators to annotate other roles not covered by our ontology as Miscellaneous as well.}

\section{Data and Annotation}\label{sec:annotation}

\subsection{Source Datasets}
We randomly sample examples from three LFQA datasets and filter answers with more than 15 sentences and those with less than 3 sentences.\footnote{We used Stanza~\cite{qi2020stanza} to split long-form answers into sentences. This process removes 42\%, 28\% and 34\% from ELI5, WebGPT and NQ respectively.} We briefly describe each dataset below.\footnote{Our data is sourced from the validation split of ELI5 from the KILT \cite{petroni-etal-2021-kilt} benchmark, the testing portion from WebGPT (their samples are publicly hosted at \url{https://openaipublic.blob.core.windows.net/webgpt-answer-viewer/index.html}, which answers questions from the ELI5 test set), and the validation split from Natural Questions.}


\paragraph{ELI5 / ELI5-model}
ELI5 consists of QA pairs where the questions and answers are retrieved from the subreddit \texttt{r/explainlikeimfive}. 
The answers in ELI5 are of varying quality and style. While the original dataset consists of (question, answer) pairs, recent benchmark~\cite{petroni-etal-2021-kilt} annotated a subset of examples with relevant Wikipedia paragraphs, which we used for analysis in Section \ref{sec:discourse_analysis}. In addition to answers in the original datasets, we annotate a small number of model-generated answers from \citet{Krishna2021HurdlesTP} (we refer this set as ELI5-model), a state-of-the art LFQA system on ELI5.


\paragraph{WebGPT}
\citet{nakano2021webgpt} presented a new LFQA dataset and model;
with the goal of building a model that can search and navigate the web to compose a long-form answer. While they re-use questions from ELI5, they newly collect answers from trained human annotators who were instructed to first search for related documents using a search engine and then construct the answers with reference to those documents. The collected data (denoted as ``human demonstration'' consisting of question, answer, a set of evidence documents, and mapping from the answer to the evidence document) are used to finetune GPT-3~\cite{NEURIPS2020_1457c0d6} to generate long-form answers.

\paragraph{Natural Questions (NQ)}
NQ contains questions from Google search queries, which is paired with a relevant Wikipedia article and an answer in the article if the article answers the question. They annotate paragraph-level answer as well as short span answer inside the paragraph answer if it exists. In open retrieval QA, researchers~\cite{orqa} filtered questions with paragraph level answers for its difficulty of evaluation and only look at questions with short span answer. 

We create a filtered set of NQ that focuses on paragraph-level answers containing complex queries.\footnote{Implementation details are in \ref{subsec:append_imp_details}. We also release these questions in our github repository.} While many NQ questions can be answered with a short entity (e.g., \emph{how many episodes in season 2 breaking bad?}), many others questions require paragraph length answer (e.g., \emph{what does the word china mean in chinese?}). This provides a complementary view compared to the other two datasets, as the answers are not \textit{written specifically} for the questions but harvested from pre-written Wikipedia paragraphs. Thus, this simulates scenarios where model retrieves paragraphs instead of generating them.

\subsection{Annotation Process}
We have a two-stage annotation process: annotators first determine the validity of the QA pair, and proceed to discourse annotation only if they consider the QA pair valid. We define the QA pair as valid if (1) the question is interpretable, (2) the question does not have presuppositions rejected by the answer, (3) the question does not contain more than one sub-question, and (4) the proposed answer properly addresses the question. Examples of the invalid QA pair identified are in \ref{subsec:append_invalid}.\footnote{The categories are not mutually exclusive, and we let annotators to pick any of them when an example belongs to multiple categories.}

We collect the first stage annotation from US-based crowdsource workers on Amazon Mechanical Turk and second stage annotation from undergraduate students majoring in linguistics, who are native speakers in English.\footnote{Initially, we aimed to collect all data from crowdsourcing, but {during our pilot we found that it is challenging for crowd worker to make role assignment}.} A total of 29 crowdworker participated in our task, and six undergraduates annotated roles for a subset of QA pairs annotated as valid by crowdworkers. We first qualified and then provided training materials to both groups of annotators. The annotation guideline and interface can be found in \ref{subsec:append_guideline}. We paid crowd workers \$0.5 per example, and our undergraduate annotators \$13 / hour. More details of data collection can be found in our datasheet.


\begin{table}
\small
\begin{center}
\begin{tabular}{lrrr}
\toprule
\textbf{Data} & \textbf{Validity} & \textbf{Role} & \textbf{Length} \\
\midrule
\vspace{0.05cm}
ELI5 & 1,035 (6,575)  & 411 (2,670) & 6 (126)  \\
\vspace{0.05cm}
ELI5-model & 193 (1,839) & 115 (1,080) & 10 (210) \\
\vspace{0.05cm}
WebGPT & 100 (562) & 98 (551) & 6 (131) \\
\vspace{0.05cm}
NQ &  263 (1,404)  & 131 (695) & 5 (139) \\   
\midrule
\textbf{Total} & 1,591 (10,380) & 755 (4,996) & 7 (139) \\
\bottomrule
\end{tabular} 
\end{center}\vspace{-0.3em}
\caption{Data Statistics. For validity and role, the first number in each cell corresponds to the number of long-form answers, and the second number represents the number of sentences. For length, the first number corresponds to the average number of sentences and the second represents the number of words.}
\label{tab:data_stat}
\end{table}

\input{figs/avg_pairwise_cm_all}
\subsection{Data Statistics}
Table~\ref{tab:data_stat} presents the statistics of our annotated data. We collected validity annotations for 1.5K examples and role annotations for about half of them. As our tasks are complex and somewhat subjective, we collected three way annotations. 
We consider a QA pair valid if all annotated it as valid, and invalid if more than two annotated it as invalid. If two annotators considered valid, we collect one additional annotation and consider it valid if and only if the additional annotator marked it as valid.\footnote{The Fleiss kappa for agreement improves to 0.70 after this re-annotation process.} 
We consider the majority role (i.e. chosen by two or more than two annotators) as the gold label. When all annotators chose different roles, they resolved the disagreement through adjudication. We report inter-annotator agreement before the adjudication.

\paragraph{Inter-annotator Agreement}\label{subsec:iaa}
We find modest to high agreement for both annotation tasks: For crowdworkers, Fleiss Kappa was 0.51 for validity annotation. For student annotators, Fleiss Kappa was 0.44 for role annotation. Figure \ref{fig:human_role_cm} shows the confusion matrix between pairs of annotations, with the numbers normalized by row and averaged across pairs of annotators. {We observe frequent confusion between roles denoting different levels of information salience \textemdash  Answer vs.\ Answer-Summary, and Answer vs.\ Auxiliary Information, reflecting the nuance and subjectivity in judging what information is necessary to answer a complicated question. Examples can be found in  \ref{subsec:example_role_annotation}.}

\begin{table}
\centering \small
\begin{tabular}{m{3cm}rrr}
\toprule\vspace{-0.3em}
\textbf{Reason} & \ \textbf{NQ} &\textbf{ELI5} & \textbf{WebGPT} \\ \midrule
No valid answer&15\% &10\% & 1\% \\ 
Nonsensical question & 3\% & 1\% &  0\% \\
Multiple questions & 9\% & 4\% &   1\% \\
Rejected presupposition & 2\% & 10\% &  0\% \\
\midrule
\textbf{Total} & 23\% & 19\% & 2\% \\
\bottomrule
\end{tabular}
\caption{Different reasons for invalid question answer pairs and their frequency in the three datasets.}
\label{tab:invalid_qa_definition}
\end{table}

\begin{table*}
\small
\begin{center}
\begin{tabular}{lrrrrrrr}
\toprule
\multirow{2}{*}{\textbf{Data}}& \multirow{2}{*}{\textbf{\# of Annotated Sentences}}&  \multicolumn{6}{c}{\textbf{Role}}\\
&& \textbf{Answer} & \textbf{Summary} & \textbf{Auxiliary} & \textbf{Example} & \textbf{Org} & \textbf{Misc} \\ \midrule
ELI5 & 2670 &  30\% & 28\% & 18\% & 13\%& 1\%& 10\%  \\
WebGPT & 551 & 28\% & 35\% & 26\% & 8\% & 3\% & 0\% \\
NQ & 695 & 21\% & 35\% & 39\% &5\%&0.4\%& 0.1\% \\  
\midrule
Total & 3916 & 28\% & 30\% & 11\% & 23\% &1\% & 7\% \\ 
\bottomrule
\end{tabular} 
\end{center}\vspace{-0.5em}
\caption{Sentence-level role distribution. The first column represent the total number of the annotated answer sentences. The remaining column represents the proportion of each role in respective datasets.}
\label{tab:answer_type_stat}
\end{table*}

\section{Discourse Analysis of Long-form Answers}\label{sec:discourse_analysis}
With our annotated data, we study the differences between the three types of long-form answers, namely answers provided by users in online community (ELI5), answers written by trained annotators through web search (WebGPT), and answers identified in Wikipedia passages (NQ).

{\paragraph{Q/A Validity} Table \ref{tab:invalid_qa_definition} summarizes the portion of valid answers in the three datasets and the distribution of invalid reasons. NQ has the highest rate of invalid answer (15\%). Upon manual inspection, we find that passages from Wikipedia written independently of the question often only partially address complex questions. This demonstrates the limitation of a fully extractive approach. Around 10\% of the answers from ELI5 reject presupposition in the question, which is a common phenomena in information-seeking questions~\cite{kim2021linguist}. WebGPT boasts the lowest invalid rate, showing the high quality of their collected answers.}

{\paragraph{Role Distribution} We study the distribution of roles in three datasets (Table \ref{tab:answer_type_stat}). NQ shows the highest proportion of auxiliary information, as the paragraphs are written independent of the questions. In contrast, ELI5 contains more answer sentences and examples which provide explanation. Both ELI5 and WebGPT contain organizational sentences, demonstrating that it is commonly used when answerers assemble answers that cover more than one aspects. In all datasets, around half of the sentences serve roles other than directly answering the questions, such as providing auxiliary information or giving an example, which reflects the wide spectrum of information presented in a long-form answer. Relatively few sentences (less than 10\%) are marked as miscellaneous, showing a high coverage of our ontology in the three LFQA datasets we investigated.} Compared to ELI5, both WebGPT and NQ answers contain very little miscellaneous sentences. This is partially because both datasets are more extractive and less personal, without sentences which serve the role of various kinds of communication from answerers to question askers (e.g. expressing sentiments, pointing to other resources) that are commonly seen in online community forum.

{\paragraph{Discourse Structure} Figure~\ref{fig:role_distribution_normed} presents the distribution of each role per its relative location in the answer. Despite the significant differences in the proportion of different discourse roles, the positioning of the roles is similar across the datasets. Answer summary and organizational sentences typically locate at the beginning of the paragraph, examples and answers often in the middle, with an increasing portion of auxiliary information towards the end. The sentences belonging to miscellaneous role frequently position at the beginning or the end of the paragraph, instead of intervening in the middle. WebGPT contains a higher portion of auxiliary information locating at the beginning of the passage, followed by the answer summary sentences.}

\input{figs/role_idx_perc_merged_normed}

\paragraph{Answer Extractiveness}
One important aspect for long-form answer is whether the answer can be attributed to an external evidence document. While answers from NQ are directly extracted from Wikipedia passages, both ELI5 and WebGPT are written specifically for the question. To help with verification, both datasets provide evidence documents paired with the answer, and yet there are design differences between the two. Answerer (annotators) of WebGPT were instructed to answer the question \textit{based on} the evidence documents returned by a search engine, while answers from ELI5 were written first \textit{independently} and later paired with relevant Wikipedia passages \cite{petroni-etal-2021-kilt}.

We found that such difference leads to different level of extractiveness of the answer, by calculating sentence-level lexical overlap (after removing stopwords) with the evidence document. Overall,  WebGPT answers exhibit \textbf{more} lexical overlap (unigram: 0.64, bigram: 0.36) with evidence document than ELI5 answers (unigram: 0.09, bigram: 0.01). Answer sentences with different roles also exhibit different levels of extractiveness (detailed role-level overlap can be found in Table \ref{tab:answer_ref_overlap} in the appendix). For ELI5 answers, sentences belonging to answer and summary roles have the highest overlap while example, auxiliary information and miscellaneous sentences are less \textit{grounded} to external sources. For WebGPT, organizational sentences are the least extractive among all the roles.

\section{Discourse Structure of Model-generated Answers}\label{subsec:gen_answer}

Having analyzed discourse roles of human-written long-form answers, we investigate the discourse structure of model-generated answers. This will allow us to quantitatively study the difference in terms of discourse structure across gold and generated answers, which we hope will cast insights to the linguistic quality of system outputs. 

\paragraph{Systems} We study model-generated answers from a state-of-the-art LFQA system~\cite{Krishna2021HurdlesTP}.\footnote{We sampled from four different model configurations reported in their paper, i.e. combination of nucleus sampling threshold p=\{0.6, 0.9\}, and generation conditioning on \{predicted, random\} passages. The answers we annotated achieved a ROUGE-L of 23.19, higher than that of human-written answers on the same set of questions (21.28).}  Their model uses passage retriever~\cite{realm}, and generates answers based on the retrieved passage with a routing transformer model~\cite{Roy2021EfficientCS}.


\paragraph{Q/A Validity} We collect validity annotation on 193 model-generated answers, and 78 are considered invalid, significantly higher ratio than that of human written answers (Table~\ref{tab:invalid_qa_definition}). The Fleiss's kappa of QA pair validity is 0.26 (and 0.61 after collecting one more label), substantially lower than the agreement on human written answers (0.51, 0.70) while annotated by the same set of annotators. {Detailed distribution of invalid reason annotated can be found in Table \ref{tab:invalid_qa_definition_gen}. Despite the low agreement, 60 of them are marked as ``No valid answer'' by at least two annotators. The flaw of automatic measures was also pointed out by prior work \cite{Krishna2021HurdlesTP}, which compares ROUGE between human-written and model-generated answers. Our study reiterates that the generated answers received high ROUGE score \textit{without} answering the question. }

\paragraph{Roles} We proceed to collect sentence-level role annotations on 115 valid generated long-form answers following the same annotation setup in Section \ref{sec:annotation}, and hence our annotators were not asked to evaluate the \textit{correctness} or the \textit{quality} of the answers (e.g. whether the generated example makes sense), focusing on the functional roles of sentences only. We found that the annotators \textit{disagree} substantially more as compared to the human-written answers, with a Fleiss kappa of 0.31 (vs.\ 0.45 for human-written answers), suggesting that the discourse structure of model-generated answers are \textit{less clear}, even to our trained annotators.

\begin{figure}
\begin{tikzpicture}[
/pgfplots/every axis/.style={
  width=.5\textwidth,
  height=.28\textwidth,
  font=\small,}]

\definecolor{color0}{rgb}{0.194607843137255,0.453431372549019,0.632843137254902}
\definecolor{color1}{rgb}{0.881862745098039,0.505392156862745,0.173039215686275}

\begin{axis}[
legend cell align={left},
legend style={
  fill opacity=0.8,
  draw opacity=1,
  text opacity=1,
  at={(0.85,0.91)},
  anchor=north,
  draw=white!80!black
},
tick align=outside,
tick pos=left,
unbounded coords=jump,
x grid style={white!69.0196078431373!black},
xmin=-0.5, xmax=6.5,
xtick style={color=black},
xtick={0,1,2,3,4,5,6},
xticklabel style={rotate=20.0, font=\tiny},
xticklabels={Sum,Ans,Aux,Ex,\color{red}{Disagreed},Misc,Org},
y grid style={white!69.0196078431373!black},
ymin=0, ymax=0.32,
ytick style={color=black},
xticklabel style={font=\tiny},
]
\draw[draw=none,fill=color0] (axis cs:-0.4,0) rectangle (axis cs:0,0.261780104712042);
\addlegendimage{ybar,ybar legend,draw=none,fill=color0}
\addlegendentry{human}

\draw[draw=none,fill=color0] (axis cs:0.6,0) rectangle (axis cs:1,0.261032161555722);
\draw[draw=none,fill=color0] (axis cs:1.6,0) rectangle (axis cs:2,0.167165295437547);
\draw[draw=none,fill=color0] (axis cs:2.6,0) rectangle (axis cs:3,0.119296933433059);
\draw[draw=none,fill=color0] (axis cs:3.6,0) rectangle (axis cs:4,0.0912490650710546);
\draw[draw=none,fill=color0] (axis cs:4.6,0) rectangle (axis cs:5,0.0893792071802543);
\draw[draw=none,fill=color0] (axis cs:5.6,0) rectangle (axis cs:6,0.0100972326103216);
\draw[draw=none,fill=color1] (axis cs:-2.77555756156289e-17,0) rectangle (axis cs:0.4,0.181481481481481);
\addlegendimage{ybar,ybar legend,draw=none,fill=color1}
\addlegendentry{model}

\draw[draw=none,fill=color1] (axis cs:1,0) rectangle (axis cs:1.4,0.234259259259259);
\draw[draw=none,fill=color1] (axis cs:2,0) rectangle (axis cs:2.4,0.275925925925926);
\draw[draw=none,fill=color1] (axis cs:3,0) rectangle (axis cs:3.4,0.0768518518518519);
\draw[draw=none,fill=color1] (axis cs:4,0) rectangle (axis cs:4.4,0.171296296296296);
\draw[draw=none,fill=color1] (axis cs:5,0) rectangle (axis cs:5.4,0.0555555555555556);
\draw[draw=none,fill=color1] (axis cs:6,0) rectangle (axis cs:6.4,0.00462962962962963);
\addplot [line width=1.08pt, white!26!black, forget plot]
table {%
-0.2 nan
-0.2 nan
};
\addplot [line width=1.08pt, white!26!black, forget plot]
table {%
0.8 nan
0.8 nan
};
\addplot [line width=1.08pt, white!26!black, forget plot]
table {%
1.8 nan
1.8 nan
};
\addplot [line width=1.08pt, white!26!black, forget plot]
table {%
2.8 nan
2.8 nan
};
\addplot [line width=1.08pt, white!26!black, forget plot]
table {%
3.8 nan
3.8 nan
};
\addplot [line width=1.08pt, white!26!black, forget plot]
table {%
4.8 nan
4.8 nan
};
\addplot [line width=1.08pt, white!26!black, forget plot]
table {%
5.8 nan
5.8 nan
};
\addplot [line width=1.08pt, white!26!black, forget plot]
table {%
0.2 nan
0.2 nan
};
\addplot [line width=1.08pt, white!26!black, forget plot]
table {%
1.2 nan
1.2 nan
};
\addplot [line width=1.08pt, white!26!black, forget plot]
table {%
2.2 nan
2.2 nan
};
\addplot [line width=1.08pt, white!26!black, forget plot]
table {%
3.2 nan
3.2 nan
};
\addplot [line width=1.08pt, white!26!black, forget plot]
table {%
4.2 nan
4.2 nan
};
\addplot [line width=1.08pt, white!26!black, forget plot]
table {%
5.2 nan
5.2 nan
};
\addplot [line width=1.08pt, white!26!black, forget plot]
table {%
6.2 nan
6.2 nan
};
\draw (axis cs:-0.2,0.261780104712042) node[
  scale=0.5,
  anchor=south,
  text=black,
  rotate=0.0
]{0.26};
\draw (axis cs:0.8,0.261032161555722) node[
  scale=0.5,
  anchor=south,
  text=black,
  rotate=0.0
]{0.26};
\draw (axis cs:1.8,0.167165295437547) node[
  scale=0.5,
  anchor=south,
  text=black,
  rotate=0.0
]{0.17};
\draw (axis cs:2.8,0.119296933433059) node[
  scale=0.5,
  anchor=south,
  text=black,
  rotate=0.0
]{0.12};
\draw (axis cs:3.8,0.0912490650710546) node[
  scale=0.5,
  anchor=south,
  text=black,
  rotate=0.0
]{0.09};
\draw (axis cs:4.8,0.0893792071802543) node[
  scale=0.5,
  anchor=south,
  text=black,
  rotate=0.0
]{0.09};
\draw (axis cs:5.8,0.0100972326103216) node[
  scale=0.5,
  anchor=south,
  text=black,
  rotate=0.0
]{0.01};
\draw (axis cs:0.2,0.181481481481481) node[
  scale=0.5,
  anchor=south,
  text=black,
  rotate=0.0
]{0.18};
\draw (axis cs:1.2,0.234259259259259) node[
  scale=0.5,
  anchor=south,
  text=black,
  rotate=0.0
]{0.23};
\draw (axis cs:2.2,0.275925925925926) node[
  scale=0.5,
  anchor=south,
  text=black,
  rotate=0.0
]{0.28};
\draw (axis cs:3.2,0.0768518518518519) node[
  scale=0.5,
  anchor=south,
  text=black,
  rotate=0.0
]{0.08};
\draw (axis cs:4.2,0.171296296296296) node[
  scale=0.5,
  anchor=south,
  text=black,
  rotate=0.0
]{0.17};
\draw (axis cs:5.2,0.0555555555555556) node[
  scale=0.5,
  anchor=south,
  text=black,
  rotate=0.0
]{0.06};
\draw (axis cs:6.2,0.00462962962962963) node[
  scale=0.5,
  anchor=south,
  text=black,
  rotate=0.0
]{0.00};
\end{axis}
\end{tikzpicture}\vspace{-0.8em}
\caption{Annotated role distribution of model generated v.s. human written answers for ELI5 dataset, denoted by \% sentence. {We plot the majority role before adjudication and include those without a a majority role as "Disagreed".}}\vspace{-0.25em}
\label{fig:human_model_answer_role_dist}
\end{figure}
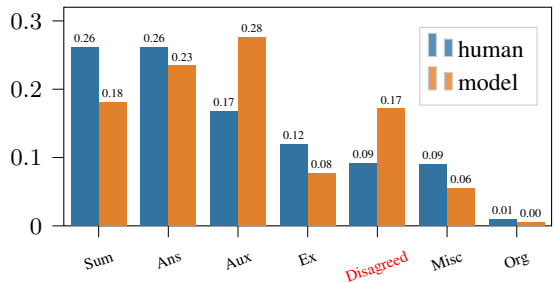

The answer role distribution of model-generated answers is very different from that of the human written answers (Figure \ref{fig:human_model_answer_role_dist}). The generated answers contain more sentences which provide auxiliary information, and fewer summary sentences. This suggests that model-generated answers contain a higher portion of information tangentially related to what is asked in the question.  Model-generated answers also contain fewer example and miscellaneous sentences.  {Examples of annotated model-generated answer can be found in Table \ref{tab:gen_role_example}.}


Overall, our results suggest that machine-generated long form answers are different from human-written answers, and judging their discourse structure is nontrivial for human annotators, resulting in lower agreement. Recent study~\cite{Karpinska2021ThePO} also showed that expert annotators showed lower agreement and took longer time to evaluate the coherence of story generated from large-scale language model.

\section{Automatic Discourse Analysis}\label{subsec:role_identification}
We study how models can identify the discourse role for each sentence in long-form answer in a valid QA pair.\footnote{We do not automatically classify QA pair validity, which requires in-depth world knowledge and goes beyond the scope of our study.}  Such a model can be beneficial for large-scale automatic analysis.

\subsection{Experimental Settings}\label{subsec:exp_setup}
\paragraph{Task and Data} Given a question $q$ and its long-form answer consisting of sentences $s_{1}, s_{2}...s_{n}$, the goal is to assign each answer sentence $s_{i}$ one of the six roles defined in Section \ref{sec:role_ontology}. As we have annotated more examples from ELI5 dataset (411 answer paragraphs compared to around 100 paragraphs in other three datasets (WebGPT, NQ and ELI5-model)), we randomly split the ELI5 long-form answers into train, validation and test sets with a 70\%/15\%/15\% ratio, and train the model on the training portion. We use all other annotated datasets for evaluation only. For model-generated answers, we filtered 185 out of 1080 sentences where model-generated sentences do not have a majority role. This setup also allows us to study domain transfer of role classification model.

\begin{table*}
\footnotesize
\centering
\begin{tabular}{lrrrrrrrrr}
\toprule
\textbf{System} & \ \textbf{Acc}  & \textbf{Match}   &  \textbf{Macro-F1} &\textbf{Ans} & \textbf{Sum} & \textbf{Aux} & \textbf{Ex} & \textbf{Org} & \textbf{Msc} \\ \midrule
Majority & 0.29 & 0.44 & 0.07  & 0 & 0.45 & 0 & 0 & 0 & 0 \\
Summary-lead & 0.36 & 0.56 & 0.15 & 0.44 & 0.46 & 0 & 0 & 0 & 0\\ 
RoBERTa & 0.46 & 0.65 & 0.43  &0.41&0.54&0.31&0.43&0.22&0.61\\ 
T5-base & 0.48 & 0.67 & 0.45 & 0.44 &0.46&0.35&0.52&0.06&0.86 \\ 
T5-large & \textbf{0.54} & \textbf{0.75} & \textbf{0.55} & \textbf{0.49} & \textbf{0.55} & \textbf{0.46} & \textbf{0.59} & \textbf{0.44} & \textbf{0.79} \\ 
\midrule
Human (l) & 0.55  & 0.73 & 0.52 &0.45 & 0.66 & 0.44 & 0.57 & 0.29 & 0.74 \\ 
Human (u) & 0.76 & 1.00 & 0.74 &0.71&0.82&0.69&0.77&0.56&0.86 \\ 
\bottomrule
\end{tabular}
\caption{Role identification results on test split of ELI5 dataset.}
\label{tab:role_identification_results_eli5_test}
\end{table*}

\begin{table*}
\footnotesize
\centering
\begin{tabular}{lccccccccc}
\toprule
\multirow{2}{*}{\textbf{System}} & \multicolumn{3}{c}{\textbf{Acc}}  & \multicolumn{3}{c}{\textbf{Match}} &  \multicolumn{3}{c}{\textbf{Macro-F1}} \\ \vspace{0.1em}
& WebGPT & NQ & ELI5-Model & WebGPT & NQ & ELI5-Model & WebGPT & NQ & ELI5-Model \\
\midrule 
 
Majority &  0.35 & 0.35 & 0.22 & 0.56 & 0.50 & 0.35 & 0.09 & 0.09 & 0.06 \\
Summary-lead &  0.35 & 0.34 & 0.35 & 0.54 & 0.56 & 0.53 & 0.15 & 0.15 & 0.16 \\
RoBERTa & 0.39& 0.45 & 0.45 & 0.60 & 0.64 &0.62   & 0.32 & 0.33 & 0.39\\
T5-base & 0.43 & 0.44 & 0.43 & 0.65 & 0.61 & 0.58 & 0.38 & 0.35 & 0.42 \\
T5-large & \textbf{0.48} & \textbf{0.45} & \textbf{0.46} & \textbf{0.70} & \textbf{0.64} & \textbf{0.64} & \textbf{0.46} & \textbf{0.40} & \textbf{0.48}\\
 \midrule
 Human (l) & 0.53 & 0.59 & 0.57  & 0.71 & 0.75 & 0.78 & 0.45 & 0.43 & 0.58 \\
 Human (u) & 0.73 & 0.78 & 0.78 & 1.00 & 1.00 & 1.00  & 0.61 & 0.66 & 0.78 \\
\bottomrule
\end{tabular}
\caption{Role identification results on out-of-domain datasets. Per-role performances are in Table \ref{tab:ood_role_metrics}. }
\label{tab:out_of_domain}
\end{table*}

\paragraph{Metrics} We report accuracy with respect to the majority role label (or adjudicated one, if majority doesn't exist) (\textbf{Acc}), match on any label from three annotators (\textbf{Match}), \textbf{F1} score for each role and their macro average score \textbf{Macro-F1}. 




\subsection{Models}\label{subsec:models} 
\paragraph{Lower bounds} We present two simple baselines to provide lower bounds: (1) {Majority}: We predict the most frequent labels in the training data: \textit{Answer-Summary}.  (2) {Summary-lead}: We predict first two sentences as \textit{Answer-Summary}, and the rest of the sentences as \textit{Answer}.
\paragraph{Classification Models} This baseline classifies each sentence independently. We use the [CLS] token from RoBERTa-Large model~\cite{Liu2019RoBERTaAR} which encodes [$question$ <q> $ans_{1}$ ... <start> $ans_{i}$ <end> ...], where $ans_{i}$ is the $i^{th}$ sentence in the answer. The training batch size is set to 64, with the initial learning rate as $5e-5$. We used AdamW optimizer and a linear learning rate schedule. We train the model for 10 epochs and report the result of the checkpoint with best validation accuracy, averaged across three random seeds.


\paragraph{Seq2Seq Models} We use two variations (base, large) of T5 model~\cite{t5}, which take the concatenation of question and answer sentences, and output the roles for each sentence sequentially. This model predicts the roles of all sentences in the answer as a single sequence. The input sequence is [$question$ $[1]$ $ans_{1}$ $[2]$ $ans_{2}$ ...], where $ans_{i}$ denotes the $i^{th}$ sentence in the answer, and the target output sequence is set to [$[1]$ $role_{1}$ $[2]$ $role_{2}$ $[3]$...], where $role_{i}$ is the corresponding role for $ans_{i}$ (e.g. ``Answer'' for the Answer role). We limit the input/output to be 512/128 tokens. For evaluating the predicted roles, we parse the output string to identify the role predicted for each sentence. We used the batch size of 16, initial learning rate of $1e-4$ with AdamW optimizer and a linear learning rate schedule. We train the model for 30 epochs and report the result of the checkpoint with the best validation accuracy, averaged across three random seeds.

\paragraph{Human performance} We provide two approximations for human performance: upperbound (\emph{u}) and lowerbound (\emph{l}). (1) {Human (\emph{u})}: We compare each individual annotator's annotation with the majority label. This inflates human performance as one's own judgement affected the majority label. (2) {Human (\emph{l})}: We compare all pairs of annotation and calculate average F1 and accuracy of all pairs. For \textbf{Match}, we compute the match for each annotation against the other two annotations.

\subsection{Results}
Table~\ref{tab:role_identification_results_eli5_test} reports the results on ELI5 test set.\footnote{Results on validation set are in Table \ref{tab:role_identification_results_eli5_val} in the appendix.} All models outperform the majority and summary-lead baselines. The sequential prediction model (T5) significantly outperform classification model (RoBERTa) which makes a prediction per sentence. The roles with lower human agreement (auxiliary, organizational sentence, answer) also exhibit low model performances, reflecting the subjectivity and ambiguity of roles for some sentences. Overall, with a moderate amount of in-domain annotated data, our best model (T5-large) can reliably classify functional roles of sentences in the long-form answers, showing comparable performances to human lower bound.

Table~\ref{tab:out_of_domain} reports the results on the three out-of-domain datasets, WebGPT, NQ and ELI5-model (model-generated answers). Human agreement numbers are comparable across all datasets (0.53-0.59 for lower bound, 0.73-0.78 for upper bound). While T5-large still exhibits the best overall performance, all learned models perform worse, partially as the role distribution has changed. Despite trained on the ELI5 dataset, role classification model also perform worse on model-generated answers (ELI5-model), echoing our observation that human annotators find it challenging to process the discourse structure of model-generated answers. Our pilot showed that training with in-domain data improved the performances consistently, but the evaluation is on a small subset (after setting apart some for training), so we do not report it here. We anticipate that automatic role classification is feasible given moderate amount of annotation for all three human-written long-form answer datasets we study.

\section{Related Work}\label{sec:related}

\paragraph{Discourse structure.} Our work is closely related to functional structures defined through content types explored in other domains; prior work has affirmed the usefulness of these structures in downstream NLP tasks. In news, \citet{choubey2020discourse} adopted \citet{van2013news}'s content schema cataloging events (e.g., main event, anecdotal), which they showed to improve the performance of event coreference resolution.
In scientific writing, content types (e.g., background, methodology) are shown to be useful for summarization~\cite{teufel2002summarizing,cohan2018discourse}, information extraction~\cite{mizuta2006zone,liakata2012automatic}, and information retrieval~\cite{kircz1991rhetorical,liddy1991discourse}. 
The discourse structure of argumentative texts (e.g., support, rebuttal)~\cite{peldszus2013argument,becker2016argumentative,stab2017parsing} has also been applied on argumentation mining. To the best of our knowledge, no prior work has studied the discourse structure of long-form answers.

\paragraph{Question Answering.} Recent work~\cite{cao-wang-2021-controllable} have investigated the ontology of questions, which includes comparison questions, verification questions, judgement questions, etc. We construct the ontology of functional roles of answer sentences. One of the roles in our ontology is summary, yielding an extractive summarization dataset. This shares motivation with a line of work studying query-focused summarization~\cite{Xu2020QueryFM}. {Concurrent to our work, \citet{Su2022ReadBG} studies improving faithfulness of long-form answer through predicting and focusing on salient information in retrieved evidence document.} Lastly, our work build up on three datasets containing long-form answers~\cite{kwiatkowski2019natural,Fan2019ELI5LF, nakano2021webgpt} and extends the analysis of long-form answers from earlier studies~\cite{Krishna2021HurdlesTP}.

\section{Conclusion}
We present a linguistically motivated study of long-form answers. We find humans employ various strategies -- introducing sentences laying out the structure of the answer, proposing hypothetical and real examples, and summarizing main points -- to organize information. Our discourse analysis characterizes three types of long-form answers and reveals deficient discourse structures of model-generated answers. Discourse analysis can be fruitful direction for evaluating long-form answers. For instance, highlighting summary sentence(s) or sentence-level discourse role could be helpful for human evaluators to dissect long-form answers, whose length has been found to be challenging for human evaluation~\cite{Krishna2021HurdlesTP}. Trained role classifier can also evaluate the discourse structure of model-generated answers. Future work can explore using sentences belonging to the summary role to design evaluation metrics that focuses on the core parts of the answer~\cite{nenkova2004evaluating}, for assessing the \textit{correctness} of generated the answer. Exploring controllable generation, such as encouraging models to provide summaries or examples, would be another exciting avenue for future work. 

\section*{Ethical Considerations}
We annotate existing, publicly available long-form question answering datasets which might contain incorrect and outdated information and societal biases. We collected annotations through crowdsourcing platform and also by recruiting undergraduate annotators at our educational institution. We paid a reasonable hourly wage (\$13/hour) to annotators and documented our data collection process with datasheet~\cite{gebru2021datasheets}. We include studies on the extractiveness of long-form answers (how much content can be grounded to evidence document) through a coarse measure of lexical overlap. This is connected to faithfulness and reducing hallucination of QA system. Our study is limited to English sources, and we hope future work can address analysis in other languages. 


\section*{Acknowledgements}

This work was partially supported by NSF grants IIS-1850153, IIS-2107524.
We thank Kalpesh Krishna and Mohit Iyyer for sharing the model predictions and human evaluation results. We would like to thank Tanya Goyal, Jiacheng Xu, Mohit Iyyer, anonymous reviewers and meta reviewer for providing constructive feedback to improve the draft. Lastly, we thank Maanasa V Darisi, Meona Khetrapal, Matthew Micyk, Misty Peng, Payton Wages, Sydney C Willett and crowd workers for their help with the complex data annotation.

\bibliography{anthology,custom}
\bibliographystyle{acl_natbib}

\newpage

\appendix

\section{Appendix}

\label{sec:appendix}


\subsection{Invalid QA}\label{subsec:append_invalid}
We provide definitions, as well as examples of each invalid QA type. 

\paragraph{No valid answer} The answer paragraph doesn't provide a valid answer to the question.
\begin{quote}
    \small
    \textbf{[Q]}: How does drinking alcohol affect your ability to lose weight?\\
    \textbf{[A]}: Alcohol itself is extremely calorically dense.Doesn't really matter whether you're drinking a light beer or shots, alcohol itself has plenty of calories. Just think of every three shots as eating a mcdouble, with even less nutritional value.
\end{quote}

\paragraph{Nonsensical question} The question is nonsensical and it is unclear what is asked.
\begin{quote}
    \small
    \textbf{[Q]}: asia vs rest of the world cricket match\\
\end{quote}

\paragraph{Multiple questions asked} More than one question are asked in the question sentence.
\begin{quote}
    \small
    \textbf{[Q]}: what is a limpet and where does it live\\
\end{quote}

\paragraph{Assumptions in the question rejected}  The answer focuses on rejecting assumptions in the question, without answering the question.
\begin{quote}
    \small
    \textbf{[Q]}: Why is it that as we get older, we are able to handle eating hotter foods\\
    \textbf{[A]}: I'm not sure I accept the premise.Children in cultures where spicy food is common, think nothing of it.My nephews had no problem eating hot peppers when they were very young because it was just a normal part of their diet.[...]
\end{quote}


\subsection{Role annotation}\label{subsec:example_role_annotation}
We include example role annotations in Table \ref{tab:role_annotation_examples} which demonstrate disagreement between Auxiliary Information and the Answer role. Sentence 2 in answer (a) was annotated as answer by most of the annotators as it elaborates on becoming a legal `next of kin' by providing a counterfactual scenario. One annotator annotated it as auxiliary as it touches upon how the decisions would be up to the parents, which goes beyond what is asked in the question. For answer (b), while most annotators think that sentence 1 is of Answer role, one annotator annotated it as Auxiliary Information which only talks about the property of purple.

\begin{table*}
\scriptsize
\begin{center}
\begin{tabular}{m{0.3cm}|m{8cm}|m{1.5cm}|m{2cm}}
\hline
\vspace{0.15cm}
\textbf{idx} & {\textbf{(a) Question: What are the benefits of marriage in the U.S.?}} & \textbf{Role}  &\textbf{ Other annotation} \\  \hline
\vspace{0.2cm}
1 & I think one of the biggest ones is that your spouse becomes your legal 'next of kin', meaning you can make medical decisions for them, own their property after they die, etc.  &  Summary &  \\ \hline
\vspace{0.15cm}
2 & If you aren't married you are not legally a part of that person's life, so any legal or medical decisions would be up to the parents of that individual. & Answer & Auxiliary \\ \hline
\vspace{0.15cm}
3 & That's why marriage equality was important a few years ago. & Auxiliary &  \\ \hline
\vspace{0.15cm}
4 & If someone was with their partner for 15 years and then suddenly dropped dead, their partner had better hope their in-laws liked them or even supported the partnership in the first place. & Example & Auxiliary \\  \hline
\vspace{0.15cm}
5 & If not, the parents could just take the house and all the money (provided the person didn't have a will). & Example & Auxiliary \\  \hline
\vspace{0.15cm}
6 & There are probably other benefits, but I think this is one of the big ones.. & Answer & Misc \\
\hline
\multicolumn{3}{c}{}\\\hline
\vspace{0.15cm}
\textbf{idx} & {\textbf{(b) Question: what is the difference of purple and violet}} &  \textbf{Role} &\textbf{Other annotation}  \\  \hline
\vspace{0.2cm}
1 & Purple is a color intermediate between blue and red . & Answer &  Auxiliary \\ \hline
\vspace{0.15cm}
2 & It is similar to violet , but unlike violet , which is a spectral color with its own wavelength on the visible spectrum of light , purple is a composite color made by combining red and blue . & Summary & \\ \hline
\vspace{0.15cm}
3 & According to surveys in Europe and the U.S. , purple is the color most often associated with royalty , magic , mystery , and piety . & Auxiliary &  \\ \hline
\vspace{0.15cm}
4 & When combined with pink , it is associated with eroticism , femininity , and seduction . & Auxiliary  & \\  \hline
\end{tabular}
\end{center}
\caption{Question paired with their paragraph level answer. Each sentence in a paragraph level answer is annotated with its role defined in Section \ref{sec:role_ontology}. We also include the other annotated role to demonstrate cases where annotators disagree with each other. (a) is from ELI5 dataset, (b) is from NQ dataset. }
\label{tab:role_annotation_examples}
\end{table*}

\begin{table}
\small \centering
\begin{tabular}{m{7cm}}
\toprule
when did the temperance movement begin in the united states \\ 
what are the ingredients in chili con carne  \\ 
is pink rock salt the same as sea salt \\ \midrule
why is muharram the first month of the islamic calendar \\
what qualifies a citizen in the han dynasty to hold a government job  \\ 
what is the difference between cheddar and american cheese  \\ 
\bottomrule
\end{tabular}\vspace{-0.3em}
\caption{Examples of NQ long questions classified as factoid (top) v.s. non-factoid (bottom).}
\label{tab:example_nq_long_questions}
\end{table}

\begin{table}
\footnotesize
\centering
\begin{tabular}{lrrrrrr}
\toprule
\multirow{2}{*}{\textbf{Role}}  & \multicolumn{2}{c}{\textbf{1-gram}}   &  \multicolumn{2}{c}{\textbf{2-gram}} & \multicolumn{2}{c}{\textbf{\# Sentences}} \\ 
 & \textbf{E} & \textbf{W} & \textbf{E} & \textbf{W} & \textbf{E} & \textbf{W} \\ \midrule
Sum & \textbf{0.10} & 0.65 & {0.01} & 0.36 & 547 & 192 \\
Ans & \textbf{0.10} & 0.61 & {0.01} & 0.32  & 571 & 154  \\
Aux & 0.08 & \textbf{0.68} & 0.00 & \textbf{0.41} & 319 & 146\\ 
Ex & 0.07 & 0.65 & 0.00 & 0.39 & 262 & 43  \\  
Misc & 0.03 & - & 0.00 & - & 199 & -   \\  
Org & 0.07 & 0.54 & 0.01 & 0.29 & 19 & 16   \\  
\bottomrule
\end{tabular}
\caption{Unigram and bigram overlap between the answer sentence and a paired evidence for \textbf{E}LI5 and \textbf{W}ebGPT per role. The last column shows the number of annotated sentences belonging to the specific role.}
\label{tab:answer_ref_overlap}
\end{table}

\subsection{Implementation Details} \label{subsec:append_imp_details}
We use \verb|pytorch-transformers| \citet{Wolf2019HuggingFacesTS} to implement our models. The hyperparameters are manually searched by the authors.
\paragraph{Question classification model} 
A difficulty in repurposing NQ is that not all questions with paragraph answers \textit{only} actually need multiple sentences. To identify complex questions, we built a simple BERT-based classifier, trained to distinguish NQ questions with short answers (i.e., less than five tokens) and ELI5 questions. We use the [CLS] token from BERT model to perform prediction. We use the original split from the ELI5 dataset, and split the NQ open's validation set into val and test set. We preprocessed the questions by converting to lowercase and exclude punctuation to remove syntactic differences between ELI5 and NQ questions. We fine-tuned the \verb|bert-base-uncased| model for 3 epochs, with an initial learning rate of $5e-5$ and batch size of 32. We use the model with the highest validation F1 as the question classifier, which achieves F1 of 0.97 and 0.94 on validation and test set respectively. We then run this classifier to select the non factoid questions from NQ questions with long-form answers, which classifies around 10\%, out of the 27,752 NQ long questions as non-factoid. Examples are in Table \ref{tab:example_nq_long_questions}.

\begin{table*}
\centering \small
\begin{tabular}{m{3cm}rrr}
\toprule\vspace{-0.3em}
\textbf{Reason} & \ \textbf{\% answer} & \textbf{Fleiss Kappa} & \textbf{Pairwise Agreement}\\ \midrule
\multicolumn{4}{c}{\textit{ELI5-model answers}} \\ \vspace{0.5em}
No valid answer& 39\% & 0.55 & 0.82\\
Nonsensical question & 1\% & 0 & 0.99 \\
Multiple questions & 6\% & 0.33 & 0.96\\
Rejected presupposition & 8\% & 0.28 & 0.95\\
\midrule 
\multicolumn{4}{c}{\textit{ELI5, WebGPT and NQ answers}} \\ \vspace{0.5em}
No valid answer& 11\% & 0.60 & 0.99 \\
Nonsensical question & 0\% & 0.67 & 0.99 \\
Multiple questions & 5\% & 0.78 & 0.99\\
Rejected presupposition & 8\% & 0.33 & 0.99\\
\bottomrule
\end{tabular}
\caption{Different reasons for invalid question answer pairs for ELI5-model and annotator agreement. We report both Fleiss kappa and pairwise agreement after reannotation. For reference, we also report agreement for human-written answers annotated.}
\label{tab:invalid_qa_definition_gen}
\end{table*}

\begin{table*}
\footnotesize
\begin{tabular}{m{12cm}|m{2cm}}
\hline
{\textbf{Question: Do animals know they're going to die?}} & \textbf{Role}   \\  \hline
I read an article about this once, I can't find it now, but I remember reading about a dog that had been put into a room with a vacuum cleaner, and it didn't notice it was sucking in air, it just started sucking in air as normal.  &  Example \\ \hline
It was pretty amazing to watch. &  \textit{Disagreed} \\ \hline
So it was just sucking in air. & Example \\ \hline
Then, the dog got out of the room and began running around the house, running into things and being hurt. & Example \\ \hline
It eventually just died of exhaustion.  &  Example  \\ \hline
So, no, they don't know.	&  Answer \\ \hline
But it is interesting to think about. &  Miscellaneous  \\ \hline
It might have just been a part of their routine, or it might have been a learned behavior, or it might have been something they did because it was the only way they could do it, and they figured it out, and it was just a part of their routine, and they thought it was cool.& Answer \\ 
\bottomrule
\end{tabular}
\caption{An example of model-generated answer with sentence-level role annotation. }
\label{tab:gen_role_example}
\end{table*}


\begin{table*}
\footnotesize
\centering
\begin{tabular}{lrrrrrrrrr}
\toprule
\textbf{System} & \ \textbf{Acc}  & \textbf{Match}   &  \textbf{Ma-F1} &\textbf{Ans} & \textbf{Sum} & \textbf{Aux} & \textbf{Ex} & \textbf{Org} & \textbf{Msc} \\ \midrule
Majority & 0.29 & 0.44 & 0.07  & 0 & 0.44 & 0 & 0 & 0 & 0 \\
Summary & 0.34 & 0.57 & 0.14 & 0.43 & 0.43 & 0 & 0 & 0 & 0\\ 
RoBERTa & 0.45 & 0.63 & 0.52  &0.38&0.57&0.31&0.54&0.67&0.56\\ 
T5-base & 0.53 & 0.74 & 0.56 & 0.49&0.54&0.33&0.64&0.56&0.76 \\ 
T5-large & \textbf{0.57} & \textbf{0.78} & \textbf{0.64} & \textbf{0.50} & \textbf{0.58} & \textbf{0.46} & \textbf{0.71} & \textbf{0.89} & \textbf{0.71} \\ 
\midrule
Human (l) & 0.57  & 0.73 & 0.57 &0.50 & 0.66 & 0.38 & 0.68 & 0.47 & 0.73 \\ 
Human (u) & 0.76 & 1.00 & 0.65 &0.72&0.82&0.67&0.83&0.69&0.85 \\ 
\bottomrule
\end{tabular}
\caption{Role identification results on validation split of ELI5 dataset.}
\label{tab:role_identification_results_eli5_val}
\end{table*}

\begin{table*}
\centering
\footnotesize
\begin{tabular}{lrrrrrr}
\toprule
\textbf{System} &\textbf{Ans} & \textbf{Sum} & \textbf{Aux} & \textbf{Ex} & \textbf{Org} & \textbf{Msc} \\ 
Majority &  0/0/0 & 0.52/0.52/0.36 & 0/0/0  & 0/0/0  & 0/0/0  & -/0/0 \\%
Summary &   0.45/0.35/0.44 & 0.4/0.53/0.51 & 0/0/0 & 0/0/0 & 0/0/0 & -/0/0 \\ 
RoBERTa & 0.40/0.19/0.42 & 0.48/0.55/0.52 & 0.20/0.46/0.43 & 0.46/0.41/0.49 & 0.08/0.00/0.17 & -/0.38/0.31 \\ 
T5-base &  0.47/0.33/0.46 & 0.45/0.52/0.48 & 0.26/0.48/0.27 & \textbf{0.55}/0.31/0.48 & 0.14/0.00/0.37 & -/0.44/0.4 \\ 
T5-large & \textbf{0.49/0.32/0.47} & \textbf{0.51/0.54/0.53} & \textbf{0.38/0.49}/0.36 & 0.51/\textbf{0.40/0.57}  & \textbf{0.43/0.06/0.49} & -/\textbf{0.58/0.43} \\ 
 \midrule
 Human (l) & 0.40/0.35/0.49 & 0.62/0.70/0.61 & 0.54/0.65/0.57 & 0.47/0.49/0.71 & 0.65/0.10/0.60 & -/0.27/0.50 \\ 
 Human (u) & 0.66/0.63/0.75 & 0.79/0.85/0.81 & 0.74/0.82/0.79 & 0.69/0.71/0.85 & 0.80/0.41/0.78 & -/0.54/0.72 \\ 
\bottomrule
\end{tabular}
\caption{Per role performance on three out-of-domain datasets. The three numbers in each cell represents performance on WebGPT, NQ, ELI5-model in order.}
\label{tab:ood_role_metrics}
\end{table*}


\subsection{Annotation Interface} \label{subsec:append_guideline}
\begin{figure*}
    \centering
    \includegraphics[width=\textwidth]{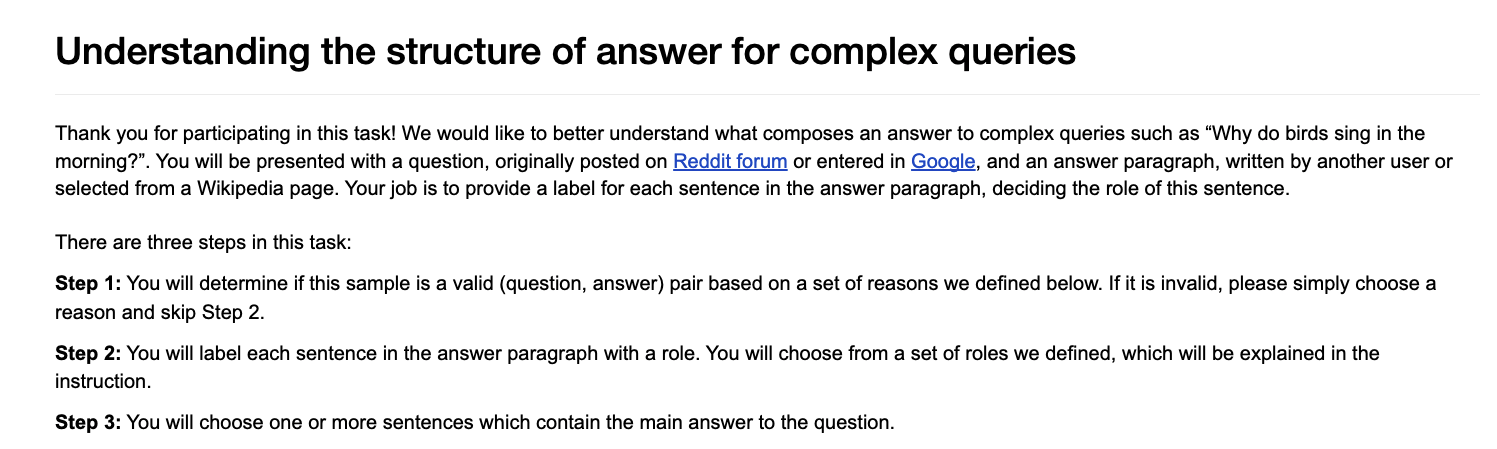}
    \caption{Screenshot of annotation guideline (overall).}
    \label{fig:guideline_overall}
\end{figure*}

\begin{figure*}
    \centering
    \includegraphics[width=\textwidth]{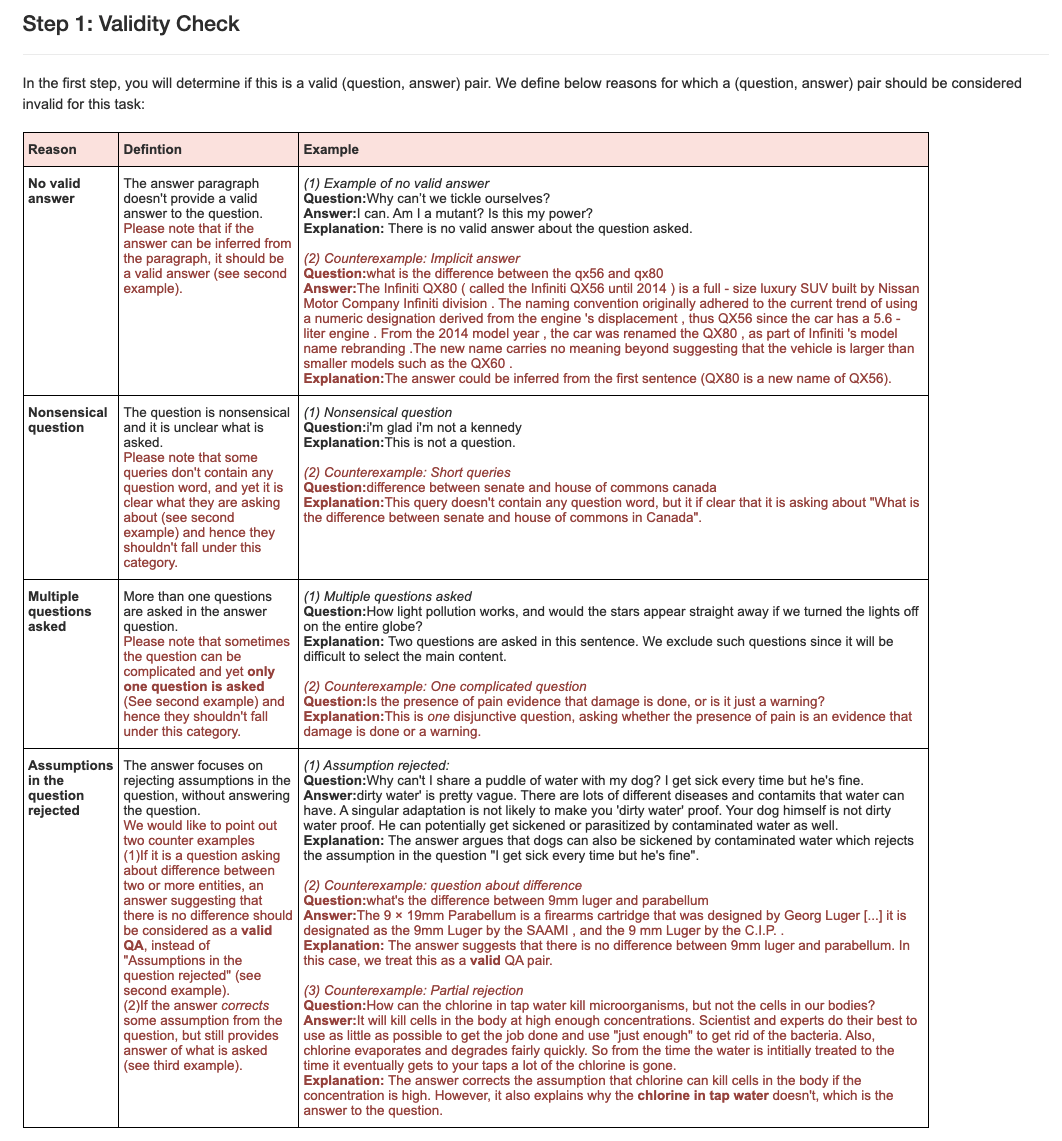}
    \caption{Screenshot of annotation guideline (Step 1).}
    \label{fig:guideline_step1}
\end{figure*}

\begin{figure*}
    \centering
    \includegraphics[width=\textwidth]{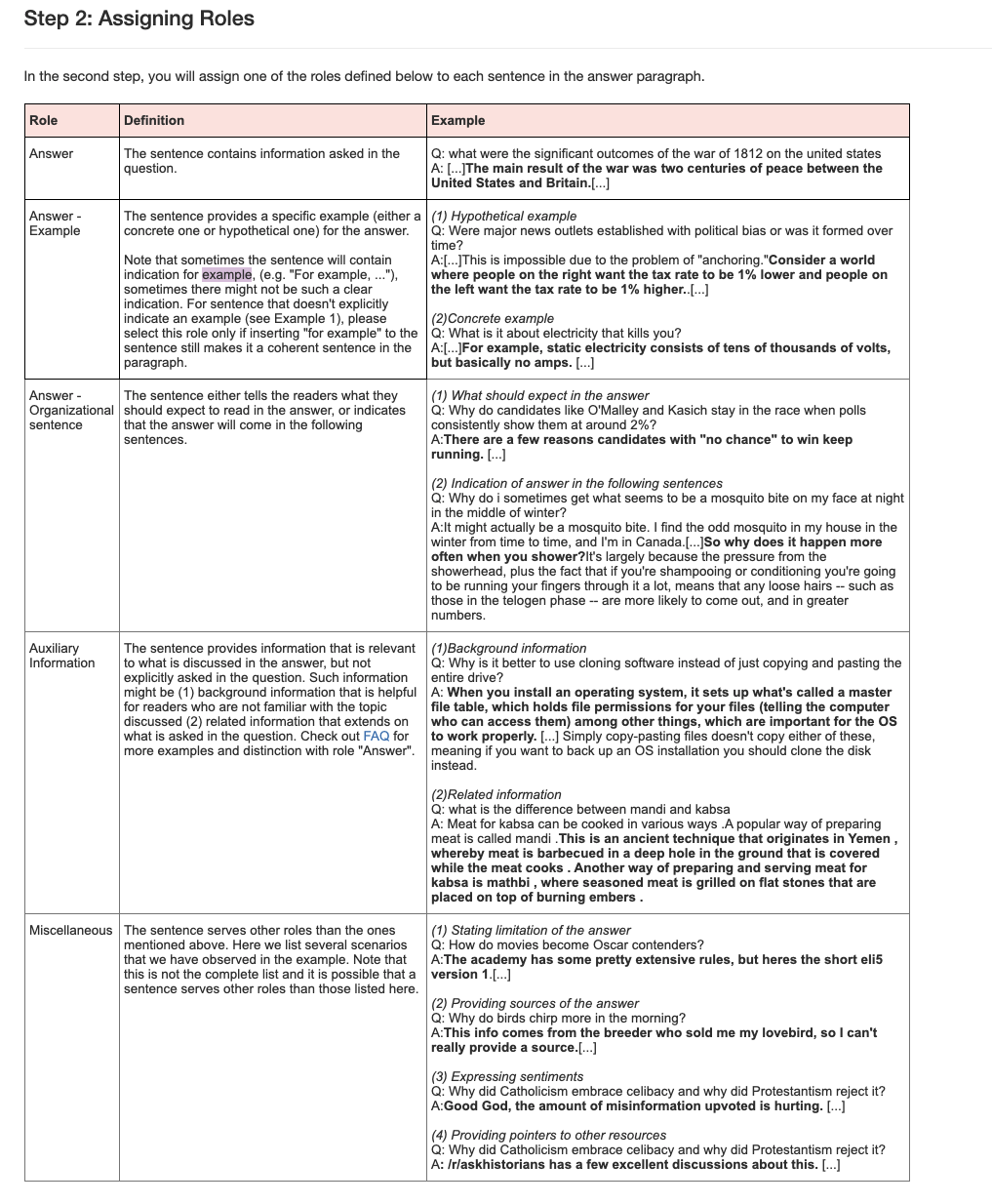}
    \caption{Screenshot of annotation guideline (Step 2).}
    \label{fig:guideline_step2}
\end{figure*}

\begin{figure*}
    \centering
    \includegraphics[width=\textwidth]{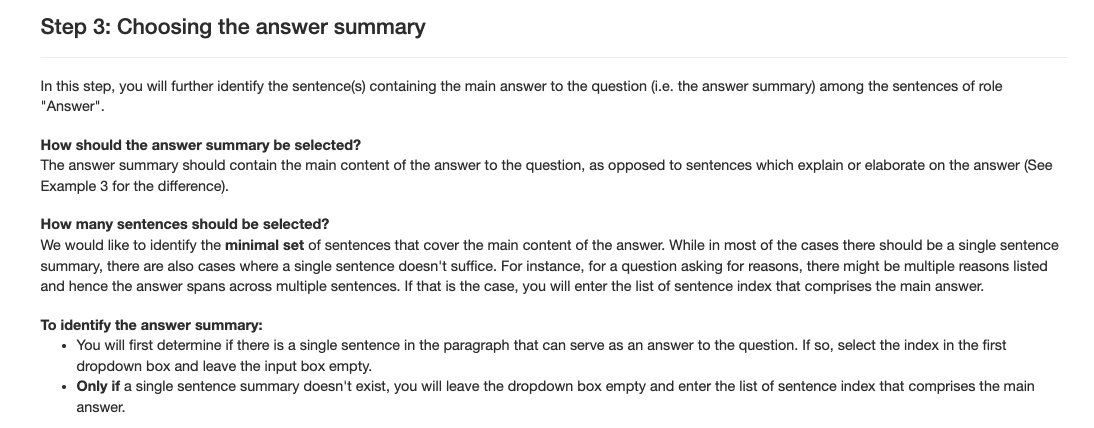}
    \caption{Screenshot of annotation guideline (Step 3).}
    \label{fig:guideline_step3}
\end{figure*}

\begin{figure*}
    \centering
    \includegraphics[width=\textwidth]{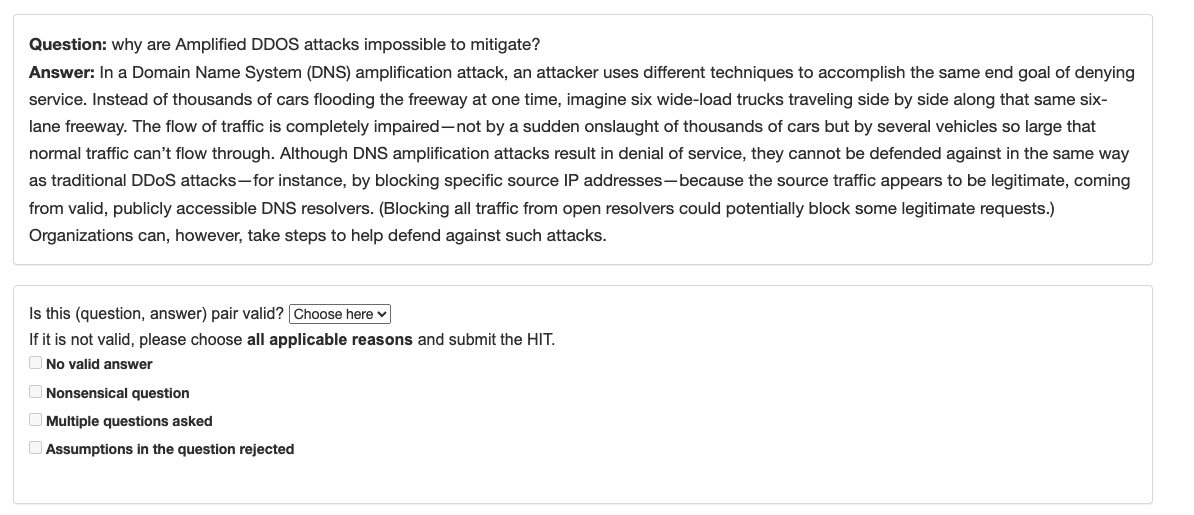}
    \caption{Screenshot of annotation interface for question validity.}
    \label{fig:step_1}
\end{figure*}

\begin{figure*}
    \centering
    \includegraphics[width=\textwidth]{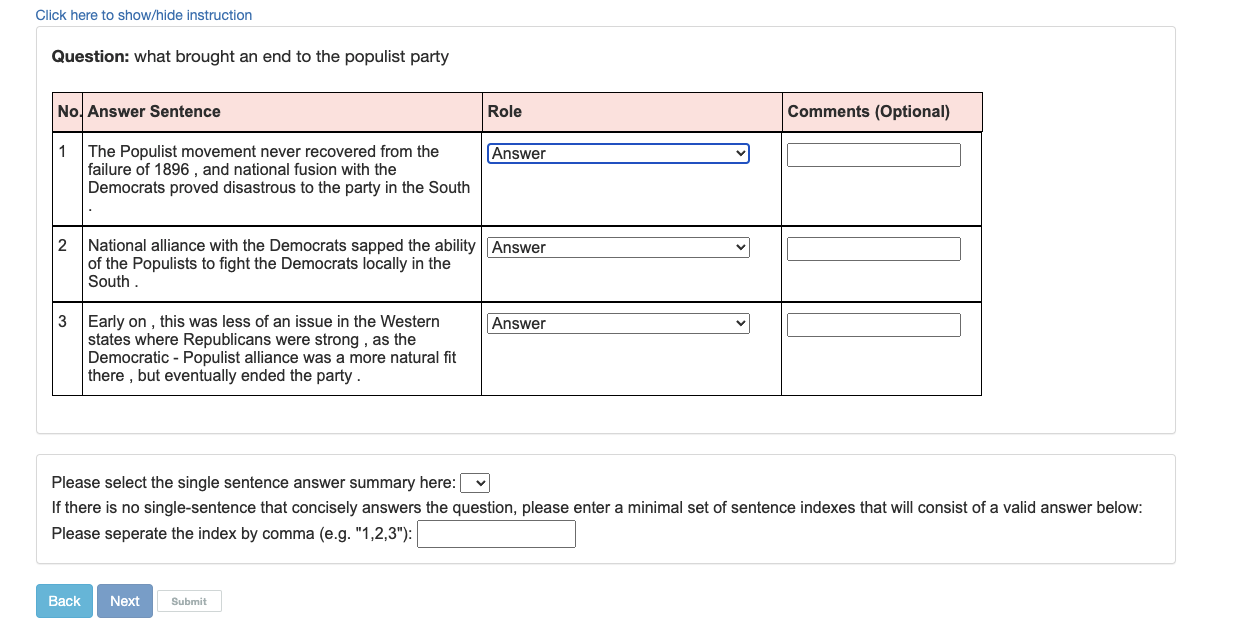}
    \caption{Screenshot of annotation interface for sentence-level role, as well as summary sentence selection.}
    \label{fig:step_2_3}
\end{figure*}

Figure \ref{fig:guideline_overall}, \ref{fig:guideline_step1}, \ref{fig:guideline_step2}, \ref{fig:guideline_step3} \ref{fig:step_1}, and  \ref{fig:step_2_3} show the annotation guideline as well as interface presented to the annotators (we present Step 1 for crowdworkers, Step 2 and Step 3 for student annotators). We didn't capture the extended example section as well as FAQ here due to space.

\end{document}